\documentclass{article}

\usepackage{arxiv}

\usepackage[utf8]{inputenc} 
\usepackage[T1]{fontenc}    
\usepackage{hyperref}       
\usepackage{url}            
\usepackage{booktabs}       
\usepackage{amsfonts}       
\usepackage{nicefrac}       
\usepackage{microtype}      
\usepackage{lipsum}
\usepackage{graphicx}
\graphicspath{ {./images/} }

\title{Automotive innovation landscaping using LLM}

\author{
 Raju Gorain \\
  Daimler Truck Innovation Center India\\
  \texttt{raju.rg.gorain@daimlertruck.com} \\
   \And
 Omkar Salunke \\
  Daimler Truck Innovation Center India \\
  \texttt{omkar.s.salunke@daimlertruck.com} \\
}

\begin{document}
\maketitle
\begin{abstract}
The process of landscaping automotive innovation through patent analysis is crucial for Research and Development teams. It aids in comprehending innovation trends, technological advancements, and the latest technologies from competitors. Traditionally, this process required intensive manual efforts. However, with the advent of Large Language Models (LLMs), it can now be automated, leading to faster and more efficient patent categorization \& state-of-the-art of inventive concept extraction. This automation can assist various R\&D teams in extracting relevant information from extensive patent databases.This paper introduces a method based on prompt engineering to extract essential information for landscaping. The information includes the problem addressed by the patent, the technology utilized, and the area of innovation within the vehicle ecosystem (such as safety, Advanced Driver Assistance Systems and more).The result demonstrates the implementation of this method to create a landscape of fuel cell technology using open-source patent data. This approach provides a comprehensive overview of the current state of fuel cell technology, offering valuable insights for future research and development in this field.
\end{abstract}


\section{Introduction}
\paragraph{}Prompt Engineering has become more popular because of its precise, well-structured way of human-AI interaction. Implementation of prompt engineering in LLM model increases the LLM output quality and accuracy. It optimizes the interaction between LLM model and user. There are many open-source transformer models which are used for extracting innovative ideas, categorization, summarization, and other NLP tasks.\\

One of the novel transformer model is BERT (Bidirectional Encoder Representations from Transformers)\cite{devlin2018bert}. BERT differs from other models by pre-training deep bidirectional representations that consider both the left and right context in all levels. By adding a single output layer, the pre-trained BERT representations may be fine-tuned to achieve cutting-edge performance in many tasks, including question answering and language inference. This can be done without making significant changes to the task-specific architecture. BERT has attained unprecedented state-of-the-art outcomes on many tasks related to natural language processing, exceeding human performance in some instances. The research\cite{devlin2018bert} emphasizes the significance of including extensive, unsupervised pre-training in language comprehension systems and applies these discoveries to deep bidirectional architectures. This model can be useful for basic categorization of patents\cite{devlin2018bert}.\\
There is a study which examines empirical scaling rules for language model performance on cross-entropy loss \cite{kaplan2020scaling}. It shows that the loss scales as a power-law with model size, dataset size, and training computation, with trends across seven orders of magnitude. Architectural features like network breadth and depth have no influence as per the research, also the paper provides simple equations for overfitting and training speed based on model/dataset size. A given compute budget may be optimally allocated using these connections. The paper finds that bigger models are more efficient, implying that ideally compute-efficient training entails training large models on little data and halting before convergence. The data strongly suggests that bigger models continue to perform better. The paper also recommends model parallelism research. So, this \cite{kaplan2020scaling} suggests the idea of using bigger pre-trained Transformer model for more accurate patent information extraction.\\
Large Language Models (LLMs) have become famous because they can do Natural Language Processing (NLP) jobs on the spot with only a few simple instructions. But users who aren't experts might find it hard to change the prompts for best performance. To fix this problem, the study presents PROMPTAID \cite{mishra2023promptaid}, a visual analytics system that lets users make, test, and improve prompts together. PROMPTAID allows to change keywords, change the way of paraphrasing, and find the best set of in-context few-shot examples. It lets users make changes to prompt templates over and over, make different questions, and see how well the created prompts work. The study says that PROMPTAID is a good tool for asking language models and suggests that more research be done on other things that affect how well prompts work in the future \cite{mishra2023promptaid}.\\
The Large Language Models (LLMs) are able to solve numerical problems. LLMs do not perform very well at solving standard logical reasoning questions from the literature of cognitive science. More tests have been conducted to see if changing the style and content of the show could make the model work better. But the changes to the presentation format and content have been found do not improve the total performance. The study has discovered that LLMs have special ways of thinking about reasoning that can't be fully predicted by human thinking and the collections of human-made words that they use. Even though standard samples have improved, LLMs can still improve in many areas of human intelligence. The study shows that LLMs' thinking doesn't match up with what humans can do. It has given the idea that better way of asking question from LLM can improve response quality, and our idea is to do it using Prompt Engineering\cite{seals2023evaluating}.\\
In one of the studies methods to extract TRIZ contradictions from patents is mentioned using openAI \cite{trapp2024llm}. The approach integrates multiple extraction steps in a single prompt and works without fine-tuning. The unique prompt technique can summarize technical contradictions, identify their parameters, and assign them to TRIZ engineering principles. In the paper, the result section shows accurate extraction of TRIZ principles using the RAG approach instead of fine-tuning which is done via GPT-4. However, the approach has potential weaknesses such as lack of reproducibility and dependency on a commercial black box LLM.\\
There is a method that utilizes Natural Language Processing (NLP) and transformer language models to extract three essential components from patents: technical challenges, solutions, and favorable consequences. The system operates at the phrase level, beyond the constraints of systems that extract information that is either too general or too particular. The model has the ability to identify the ideas, even in patents when the authors themselves neglect to label them. The system is capable of doing previous art mapping, generating innovative ideas, and identifying technical development. The study recognizes constraints associated with the used data, the formulated model, and the objective of the research. Subsequent research endeavors might enhance the golden set by conducting additional manual review conducted by domain experts. The report also proposes the need of more research in this field of study to address this deficiency \cite{giordano2023unveiling}.
\paragraph*{}
In this paper, the issue of complex readability in patent documents and the substantial time required to understand the problems and solutions outlined in these patents is addressed. Based on ideas, knowledge, expertise, and insights from previous research studies, it is believed that this issue can be resolved in a comparatively easier and less time-consuming manner. The exploration involves using a pre-trained Transformer-based Large Language Model (LLM) in conjunction with Prompt Engineering. For instance, to determine the ecosystem to which a particular patent belongs, one can obtain the answer in one or two words. In summary, this approach simplifies the process of understanding patent documents, making it more efficient and accessible.

\section{User Story \& Problem statement}
\label{sec:headings}
In automotive industry engineers always think about innovation in their products, always they have to keep update on innovations happening in their domain of work, also they have to be aware on how their competitors are handling the similar issue. For this kind of information, patents are the best resources which can be a helpful for detailed analysis of innovation. Patents contains the legal information to safeguard the rights of the inventor therefore it is considered as most authentic source of information. But as mentioned above purpose of patent text is to legal protection of the innovation and unlike the scientific research papers. For beginners the task of manual patent reading can be time consuming and exhaustive because of broad legal claims and mostly because of legal language mentioned in the document. Motivation of this paper is to extract patent information not only by keyword search but also based on the context. In this paper, patent context based information extraction is performed by Prompt Engineering technique using Transformer based LLM model.
\section{Related Work:}
This section consists of recent advancements on Prompt Engineering \& LLM model.
\subsection{Prompt Engineering}
Prompt Engineering is an effective and structured way of giving input to any LLM. Prompt can be having single question or multiple questions with content and instructions. In prompt any example can be added to extract the output in similar way. Quality of response from the LLM is depends most of the time on the structured input provided to the model. It is always advised to avoid the contradiction statements in prompt or else it might reflect the poor response.
 \begin{figure}[!tbh]
	\centering
	\includegraphics[width=135mm,height=35mm]{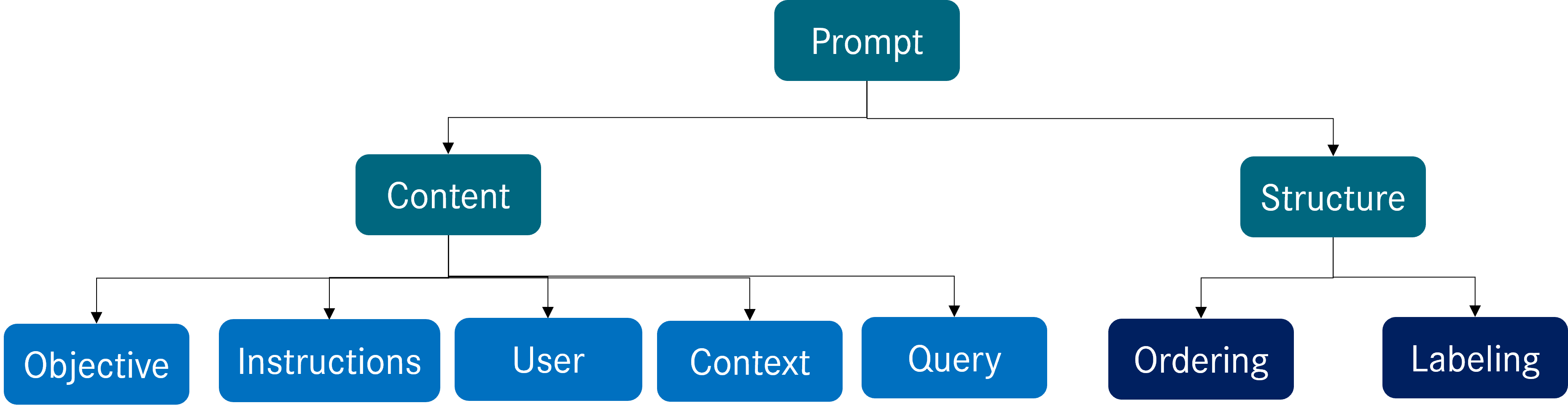}

\end{figure}

In any \textbf{effective prompt} there are two main part: \textbf{Content \& Structure}.\\
\textbf{Content} consists of \textbf{Objective}: which defines goal of the LLM or what is it’s mission, \textbf{Instruction:} is consists of the instructions on how it perform the task, \textbf{User:} it defines what role the LLM is playing , \textbf{Context:} it includes main background information/documents/input data, \textbf{Query:} it's one of the main section i.e., what information user want from LLM that need to describe in this section. \\
\textbf{Structure} consists of two main part that is Ordering and Labeling. User must keep in mind these things while creating any new prompt. \\
Below is a sample prompt:\cite{abc}
 \begin{figure}[h]
	\centering
	\includegraphics[width=85mm,height=110mm]{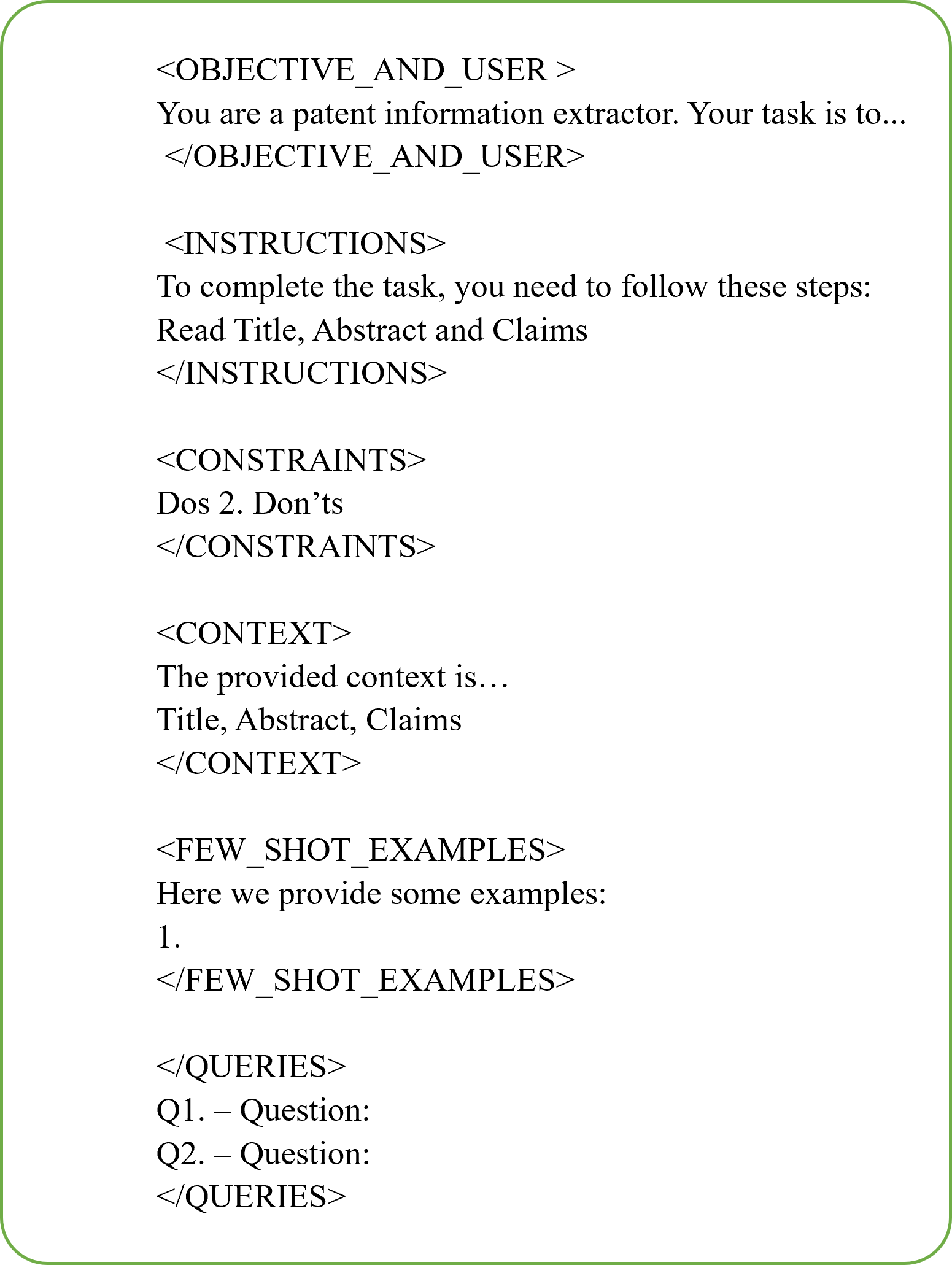}

\end{figure}

There are many techniques to generate prompts like,
\begin{itemize}
	\item \textbf{Zero-shot prompt:} Zero-shot prompting means the prompt that used to interact with the model won't contain examples or demonstrations. Zero-shot prompt directly instruct the Transformer model to perform a task without any additional examples. Sometimes output cannot be obtained if only a query is asked without context. When zero-shot doesn’t work, following few-shot can be effective.
	\item \textbf{Few shot standard prompts:} In Few shots prompting, a technique was employed to provide content by demonstrating the problem along with a instruction for better performance. Adding one example (i.e., one shot) or multiple examples (i.e., multiple shots) for demonstration can enhance performance. However, Few shots prompt is not effective for complex arithmetic, common sense, or symbolic reasoning tasks. 
\end{itemize}
And many more prompt techniques are there.

\subsection{LLM Model}
AI technologies, notably LLMs, have transformed human-AI interactions. Large Language Models are Deep Learning models that can understand human language text, can analyze with its pre-trained model, can extract required output. Most of the LLM models are trained from publicly available data of internet. LLM use a type of algorithm which can understand characters, words, sentences, and paragraphs. Deep learning uses a probabilistic approach to analysis unstructured data, which enables deep learning models to recognize the content of input data without human intervention. \\
LLM model is a pre-trained transformer model which can perform various NLP tasks like next word prediction, text generation, text classification, token generation, summarization, question answering and more. The transformer architecture is the fundamental building block of all Language Models with Transformers (LLMs). The transformer architecture was introduced in the paper “Attention is all you need,” published in December 2017  \cite{ashish2017attention}. The simplified version of the Transformer Architecture looks like this(Figure 1):
 \begin{figure}[h]
	\centering
	\includegraphics[width=150mm,height=100mm]{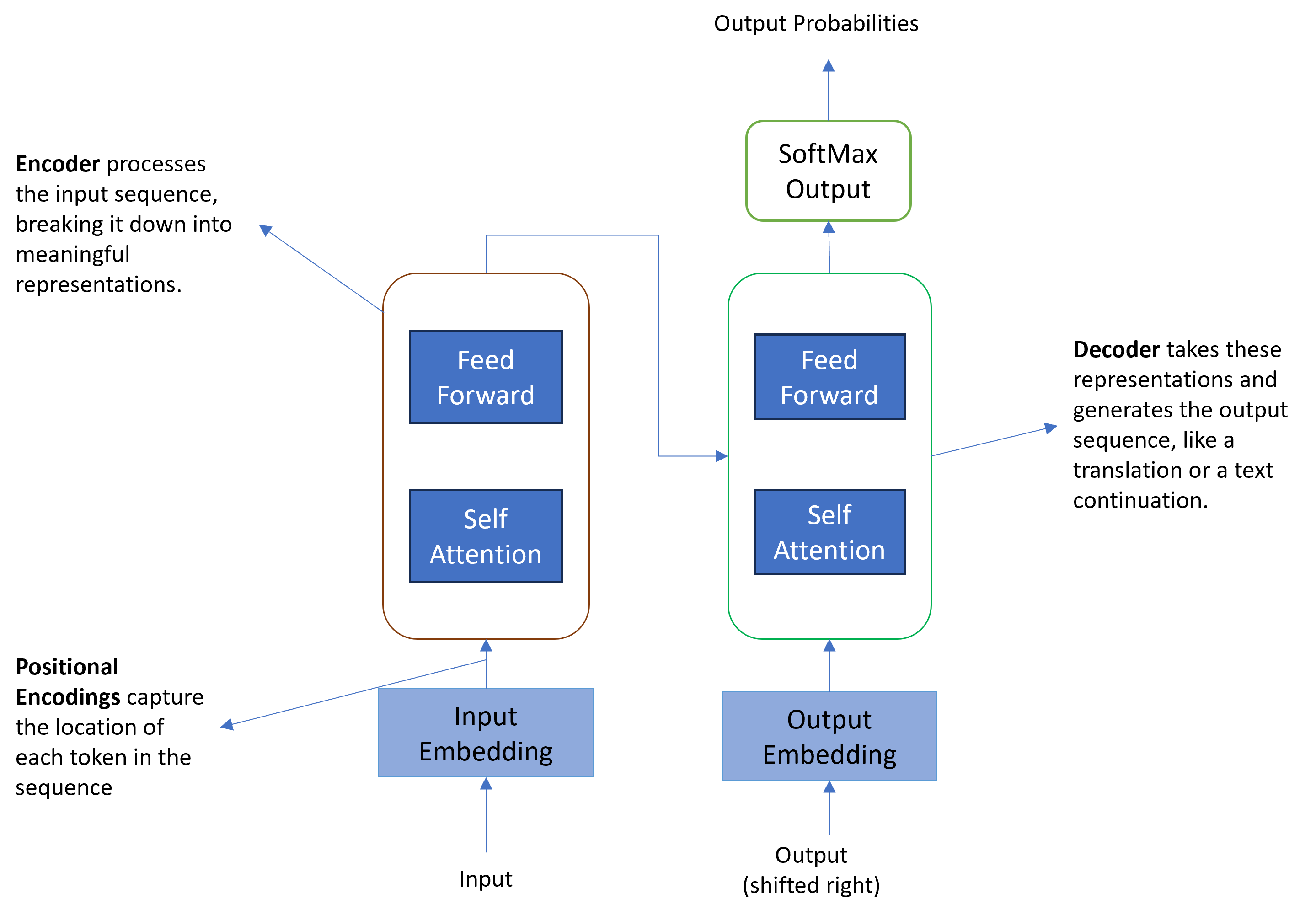}
	\caption{\small Transformer Architecture}
	
\end{figure}

\paragraph{}
LLMs are the advanced model of NLPs with more capabilities. Just like giving extra attention to important part while reading any paragraph in similar way Transformer also focuses to important word of any sentence. The transformer looks all the words at a time not one after another, this is the way it understands how words are dependent on each other. Transformer uses its knowledge to understand the story and it fits the words together. By this way it can guess what words may come next. The core of the Transformer model is nothing but deep Neural Network. All the Transformer models having different number of databases, the more accurate output can be expected from highest number of parameter model.\\
The transformer consists of two main components named, encoder and decoder. Encoder process the input sequence and break it down into model understandable representation that means it tokenize the text and decoder generate output sequence based on input sequence. Then it helps to predict the next word by looking previous words. Any transformer model consists of billions of words and sentences. Machine learning models are essentially large statistical calculators, primarily designed to process numerical data rather than text. Therefore, before feeding text into the model for processing, need to convert it into numerical tokens. This process is called embedding. Linear layer maps output embeddings into higher dimensional space, which helps to transform output- embeddings into original output input space. Then there is SoftMax function which used to generate probability distribution for each output token in the vocabulary.\\
The research in \cite{knoth2024ai} suggests that AI literacy, specifically understanding of AI technologies and human-AI cooperation, may be necessary for rapid engineering skills. AI literacy and prompt engineering can be effective to utilize generative AI tools and solve AI technology concerns. A well-structured prompt can improve this human-AI interaction which in terms helps us to extract more accurate information from LLM.
\section{Methodology}
In this paper, an application of Prompt Engineering using Transformer model is described. The objective of the paper is to extract the relevant information from patent document. A patent document consists of multiple levels of information like application number, innovators, title, abstract, claims, drawing, drawing description, summary of the patent, but this paper is considering high-level information related to patent like title, abstract and claims, which helps user to extract high-level information extraction before deep dive into patents. It has been observed that these three sections are enough to categorize patents into MATHCHEM + software categorization.
\subsection{Dataset Preparation}
This study aims to extract precise responses from a Language Model (LLM) by feeding it with well-structured text prompts. The data source for this research is patent information, which is readily accessible in an Excel format on various open-source platforms. \\
For the application use case under consideration, it requires specific patent details, namely the patent number, title, abstract, and claims. This information is obtained from a selected open-source website and subsequently extracted into an Excel format. Visual Basic for Applications (VBA) scripts have been employed to generate a prompt, which incorporates the patent content along with a question and instructions. The effectiveness of the prompt is ensured by sticking to the established rules and techniques of Prompt Engineering.\\
Upon feeding the prompt into the LLM, it generates the necessary information pertaining to the specific patent. The nature of the responses from the LLM is contingent on the questions formulated in the prompt. For a comprehensive understanding of the process, flowchart (Figure 2) is attached. This step-by-step explanation helps in easily understanding the methodology.
 \begin{figure}[h]
	\centering
	\includegraphics[width=160mm,height=55mm]{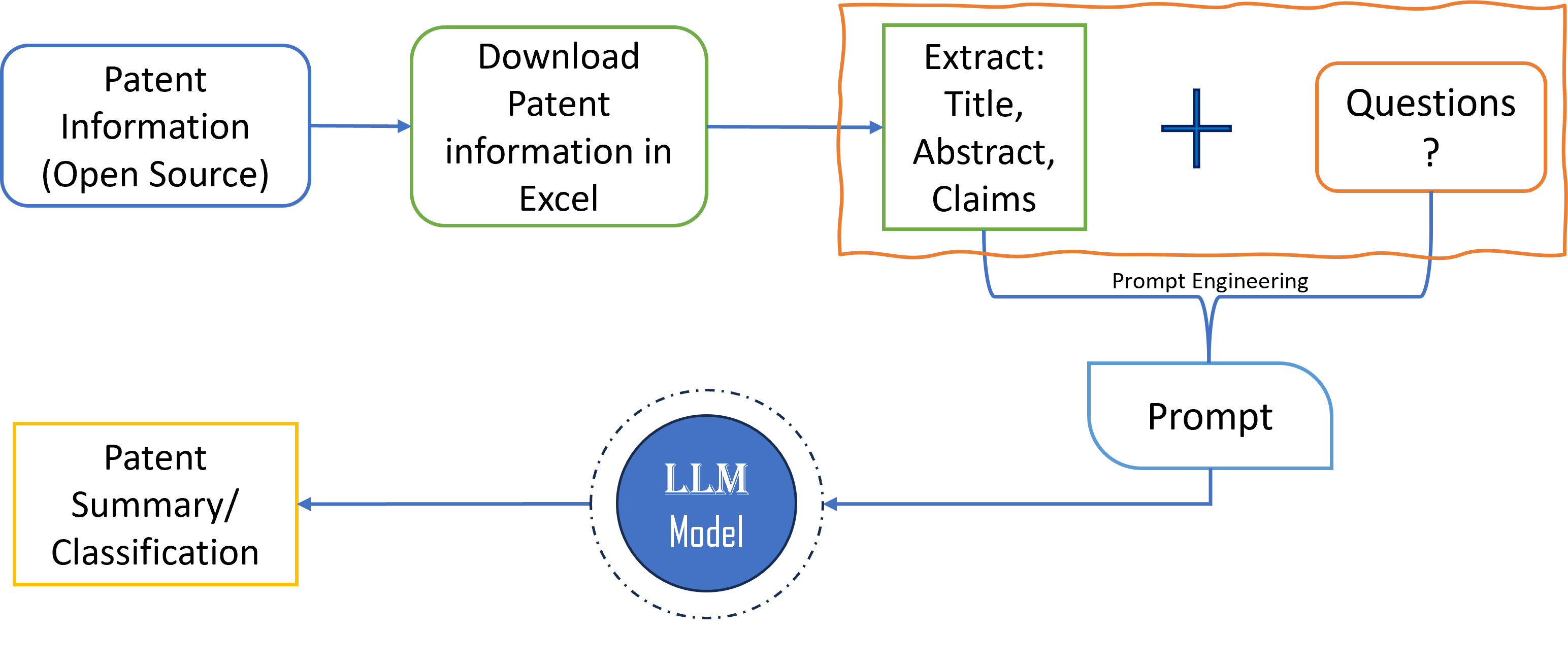}

	\caption{\small Flow Chart of Data set preparation}
\end{figure}
\subsection{Information Extraction using LLM}
As illustrated in the attached flowchart (Figure 3), various relevant pieces of information can be extracted from patent documents. Specifically, it can classify the system, comprehend the main function, identify the system components along with their features, and analyze input \& output aspects. Additionally, it provides insight into the problem, the proposed solution, and the advantages of that solution. Furthermore, it suggests how AI can solve similar problems and also hints at a potential TRIZ approach for system idealization.

 \begin{figure}[h]
	\centering
	\includegraphics[width=165mm,height=150mm]{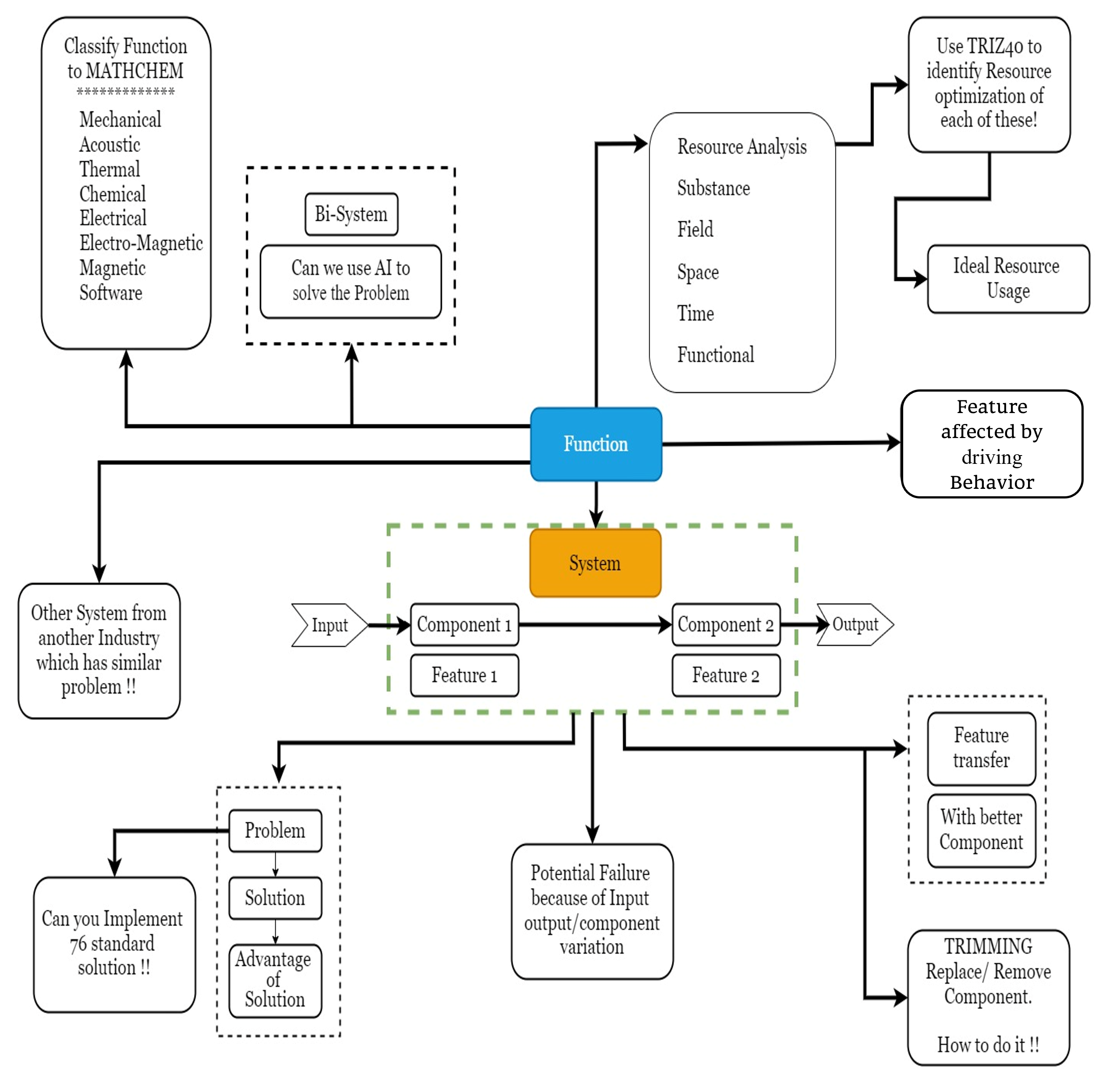}
	\caption{\small Information Extraction Flow Chart}
\end{figure}
\section{Evaluation}
Let’s see all the above-mentioned information extractions using prompt from a LLM with an example of Patent Number: US20240120514A1. Structured prompt is given to LLM, which includes title, abstract, claims of a patent document and framed questions to LLM model, and output information from LLM is extracted in given manner:\\
In the next page prompt and the LLM's result have shown,
\newpage
 \begin{figure}[h]
	\centering
	\includegraphics[width=165mm,height=75mm]{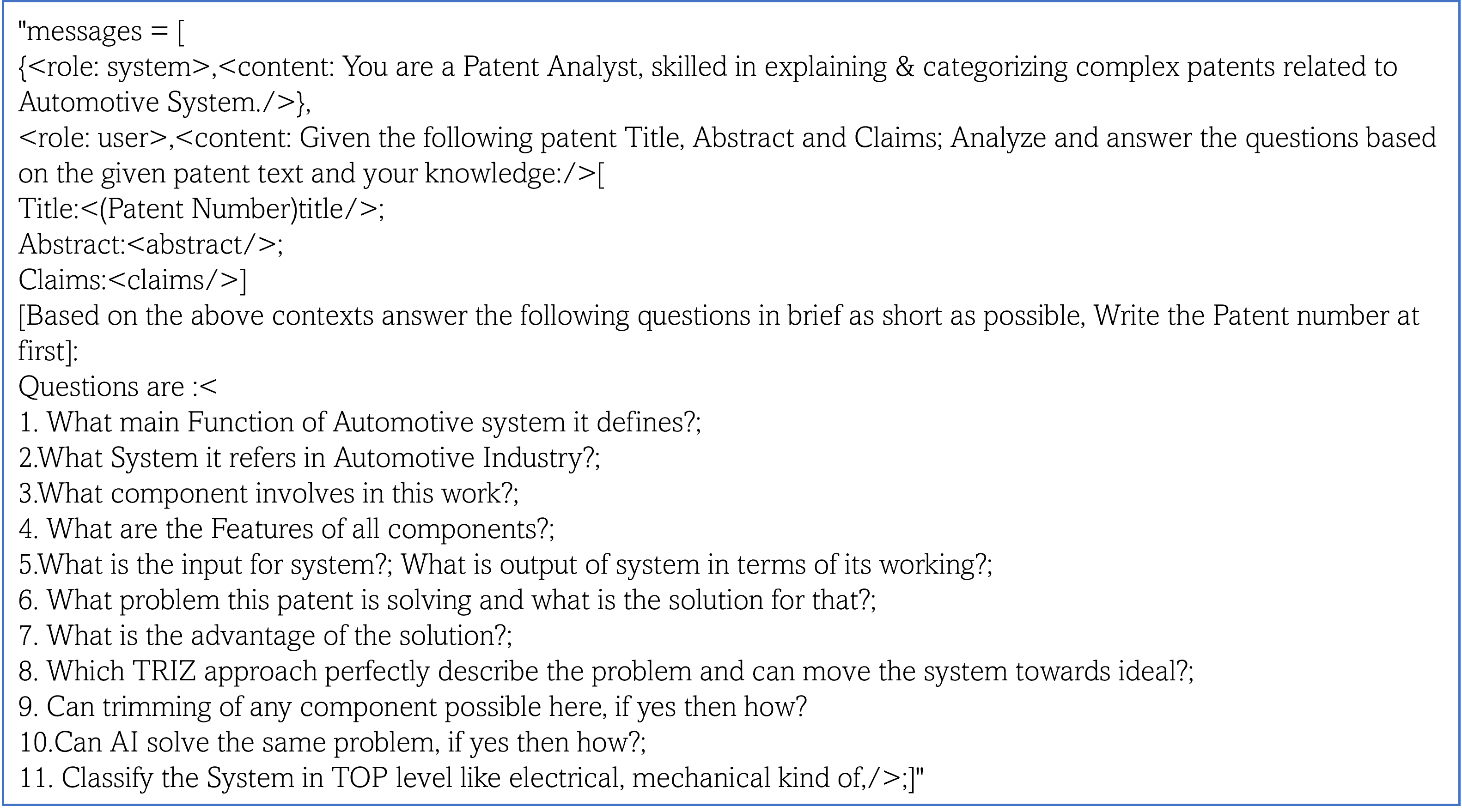}
	{\small Example Prompt}
\end{figure}
The output is:
\begin{enumerate}
	\item \textbf{System Classification:}The system can be classified at the top level as a mechanical system due to the physical components involved, and an electrical system due to the electrical processes (voltage discharge) that occur.
	\item \textbf{System in Automotive:} The system referred to in the automotive system is the fuel cell system.
	\item \textbf{Main Function:} The main function of the automotive system defined in this patent is to detect a hydrogen leak.
	\item \textbf{Components Involved:} The components involved in this work are a fuel cell stack (including a cathode and an anode), a discharge resistor, a hydrogen supply source, a system controller, and a sensor.
	\item \textbf{Features of Components:} The fuel cell stack includes a cathode and an anode with an inlet. The discharge resistor discharges voltage from the fuel cell stack. The hydrogen supply source provides hydrogen to the inlet of the anode. The system controller identifies the presence of a hydrogen leak based on measurements. The sensor measures pressure and temperature.
	\item \textbf{Input for System:} The input for the system is hydrogen supplied to the anode inlet.
	\item \textbf{Output of System:} The output is the identification of a hydrogen leak based on the rate of voltage discharge or negative pressure measurements at the anode inlet.
	\item \textbf{Problem \& Solution:} The patent solves the problem of hydrogen leaks in a fuel cell system. The solution involves initiating a shutdown process, pulsing hydrogen into the reservoir or directly into the fuel cell stack, making the system enter a discharge state, and detecting the hydrogen leak based on the rate of voltage discharge or negative pressure measurements.
	\item \textbf{Advantage of Solution:} The advantage of this solution is the ability to detect hydrogen leaks accurately and efficiently, which can prevent potential damage and improve the safety and performance of the fuel cell system.
	\item \textbf{TRIZ Approach:} The TRIZ approach that perfectly describes the problem is the “Principle of Feedback” as the system uses feedback (rate of voltage discharge or negative pressure measurements) to detect the presence of a leak.
	\item \textbf{Trimming of Component:} Trimming might be possible by integrating the functions of the sensor and the system controller into a single component, thereby reducing the complexity of the system.
	\item \textbf{AI Solution:} AI could potentially solve the same problem by using machine learning algorithms to predict the presence of a leak based on historical data and real-time measurements.
\end{enumerate}
\paragraph*{}
It is clearly visible from above mentioned example that Large Language Model can extract many relevant, important information from any patent, which not only helps to get overview of patent or filter the patents but also it gives more exposure to engineers on their innovation path. Additionally, it suggests relevant future innovations within the same domain, empowering engineers to improve their upcoming products. Engineers can get idea if it’s possible to trim or replace any component, which can help in cost reduction and system complexity reduction. In simple word, it becomes perfect assistant of R\&D engineers in their innovation journey. In the similar way by providing any other patent text as prompt anyone can extract their related information easily. This approach is not only saving time of patent reading but also making complex patent simple.
 \begin{figure}[h]
	\centering
	\includegraphics[width=160mm,height=165mm]{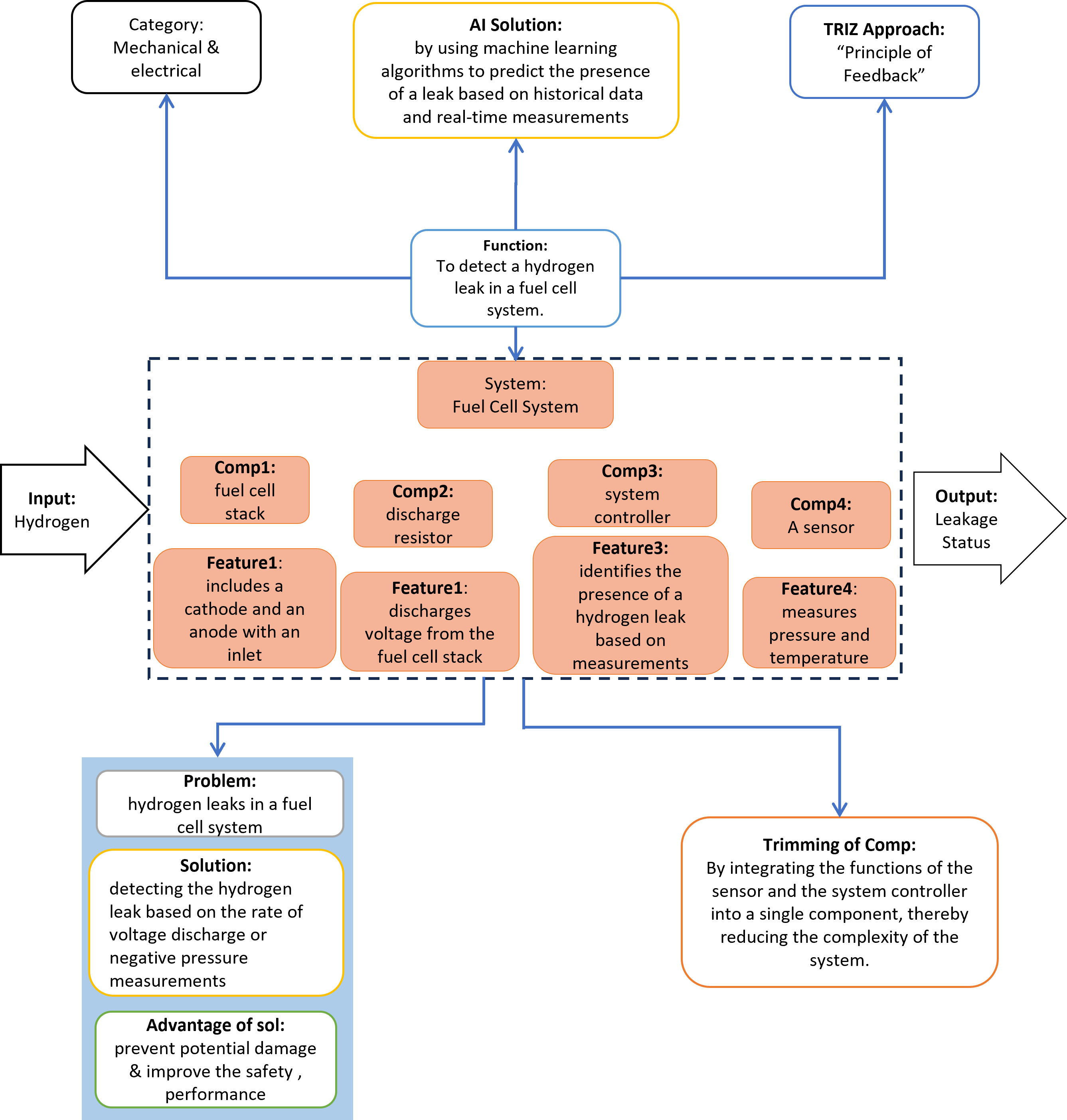}
	\caption{\small Visualization of LLM output}
	
\end{figure}

\section{Conclusion \& Future Scope}
This paper presented the method of extracting essential information from patents using prompt engineering, Approach used in the paper focuses mostly on the information extraction from the automotive powertrain innovations. In future work this approach can be extended to other areas of innovation outside powertrain domain.

\bibliographystyle{unsrt}
\bibliography{references}

\end{document}